\documentclass[lettersize,journal]{IEEEtran}
\usepackage{amsmath,amssymb,amsfonts}
\usepackage{algorithmic}
\usepackage{algorithm}
\usepackage{array}
\usepackage{textcomp}
\usepackage{url}
\usepackage{verbatim}
\usepackage{graphicx}
\usepackage{cite}

\usepackage{booktabs}
\usepackage{subcaption}
\usepackage{glossaries}
\usepackage{dirtree}

\usepackage{tikz}
\usepackage{textcomp}
\usepackage{hyperref}
\usepackage{lipsum}

\newcommand\copyrighttext{%
  \footnotesize \textcopyright 2024 IEEE. Personal use of this material is permitted. Permission from IEEE must be obtained for all other uses, in any current or future media, including reprinting/republishing this material for advertising or promotional purposes, creating new collective works, for resale or redistribution to servers or lists, or reuse of any copyrighted component of this work in other works.
  DOI: 10.1109/MITS.2024.3378114}
\newcommand\copyrightnotice{%
\begin{tikzpicture}[remember picture,overlay]
\node[anchor=south,yshift=10pt] at (current page.south) {\fbox{\parbox{\dimexpr\textwidth-\fboxsep-\fboxrule\relax}{\copyrighttext}}};
\end{tikzpicture}%
}

\begin{document}

% list of acronyms \acrshort{bla} \acrfull{bla} \gls{bla}
\newacronym{ADAS}{ADAS}{Advanced Driving Assistant Systems}
\newacronym{LKAS}{LKAS}{Lane Keeping Assistance System}
\newacronym{LKA}{LKA}{Lane Keeping Assist}
\newacronym{LDW}{LDW}{Lane departure warning}
\newacronym{LCA}{LCA}{Lane Change Assistant}

\newacronym{ALC}{ALC}{Adaptive Light Control}
\newacronym{ACC}{ACC}{Adaptive Cruise Control}
\newacronym{FCW}{FCW}{Forward Collision Warning}

\newacronym{ADS}{ADS}{Automated Driving Systems}
\newacronym{CNN}{CNN}{Convolutional Neural Network }
\newacronym{ROS}{ROS}{Robot Operating System}
\newacronym{PCL}{PCL}{Point Cloud Library}
\newacronym{SAE}{SAE}{Society of Automotive Engineers}
\newacronym{DDT}{DDT}{Dynamic Driving Task}
\newacronym{ODD}{ODD}{Operational Design Domain}
\newacronym{LDS}{LDS}{Laser Distance Sensor}
\newacronym{LIDAR}{LIDAR}{Light Detection And Ranging}
\newacronym{PCD}{PCD}{Point Cloud Data}
\newacronym{PNG}{PNG}{Portable Network Graphic}

\newacronym{IMU}{IMU}{Inertial Measurement Unit }
\newacronym{GPS}{GPS}{Global Positioning System}
\newacronym{GNSS}{GNSS}{Global Navigation Satellite System}
\newacronym{PPS}{PPS}{Pulse per Second}
\newacronym{FoV}{FoV}{Field of View}
\newacronym{Gbps}{Gbps}{Gigabits per second}
\newacronym{OBDII}{OBDII}{On-board Diagnostics II}
\newacronym{CAN}{CAN}{ Controller Area Network}
\newacronym{PoE}{PoE}{Power over Ethernet}
\newacronym{AR}{AR}{Augmented Reality}
\newacronym{SSD}{SSD}{Solid State Drive}

\title{Interaction of Autonomous and Manually Controlled Vehicles Multiscenario Vehicle Interaction Dataset}

\author{\IEEEauthorblockN{
Novel Certad\IEEEauthorrefmark{1}, \IEEEmembership{Graduate Student Member, IEEE},
Enrico del Re\IEEEauthorrefmark{1}, \IEEEmembership{Graduate Student Member, IEEE},
Helena Korndoerfer\IEEEauthorrefmark{1},
Gregory Schroeder\IEEEauthorrefmark{1}, \IEEEmembership{Member, IEEE},
Walter Morales-Alvarez\IEEEauthorrefmark{1}, \IEEEmembership{Member, IEEE},
Sebastian Tschernuth\IEEEauthorrefmark{1}, \IEEEmembership{Student Member, IEEE},
Delgermaa Gankhuyag\IEEEauthorrefmark{1},
Luigi del Re\IEEEauthorrefmark{2}, \IEEEmembership{Senior Member, IEEE},
Cristina Olaverri-Monreal\IEEEauthorrefmark{1}, \IEEEmembership{Senior Member, IEEE}
}
\\
\IEEEauthorblockA{
\IEEEauthorrefmark{1} Department Intelligent Transport Systems
\\Johannes Kepler University Linz, 4040 Linz, Austria\\
\IEEEauthorrefmark{2}Institute for Design and Control of Mechatronical Systems
\\Johannes Kepler University Linz, 4040 Linz, Austria
}

        % <-this % stops a space
\thanks{This work was supported by the Austrian Science Fund (FWF), project number P 34485-N. It was additionally supported  by the Austrian Ministry for Climate Action, Environment, Energy, Mobility, Innovation, and Technology (BMK) Endowed Professorship for Sustainable Transport Logistics 4.0., IAV France S.A.S.U., IAV GmbH, Austrian Post AG and the UAS Technikum Wien.}% <-this % stops a space
% \thanks{Manuscript received April 19, 2021; revised August 16, 2023.}
}

% \markboth{IEEE Intelligent Transportation Systems Magazine}%
% {Certad \MakeLowercase{\textit{et al.}}: IAMCV Dataset}

% \IEEEpubid{0000--0000/00\$00.00~\copyright~2023 IEEE}

\maketitle
\copyrightnotice
\begin{abstract}
The acquisition and analysis of high-quality sensor data plays a pivotal role in the development of advanced autonomous driving systems and in enhancing transportation safety. In this study, we presented a novel and extensive dataset centered on inter-vehicle interactions. The dataset was recorded across diverse locations in Germany, including roundabouts, intersections, country roads, and highways. Equipped with a sophisticated array of sensors, including LIDAR sensors, cameras, IMU/GPS, and acquisition of the vehicle bus data, our dataset offers a comprehensive representation of real-world driving scenarios. Additionally, two proof-of-concept use cases were implemented to demonstrate the versatility of the IAMCV dataset. First, an unsupervised trajectory clustering algorithm was used to highlight the capability of the dataset in categorizing vehicle movements without labeled training data. Finally, an online camera calibration method and the ROS-based standard were compared using the images captured in the dataset.
\end{abstract}


\section{Introduction}
\label{sec:introduction}
\IEEEPARstart{T}{he} rapid evolution of autonomous driving technology necessitates the availability of robust datasets that accurately reflect the complexities of real-world driving environments. With continuous advancements in sensor technologies, data acquisition systems have become more sophisticated, enabling the collection of detailed information encompassing the spatial, temporal, and contextual aspects of driving scenarios. In this context, we present a comprehensive dataset recorded using a state-of-the-art vehicle equipped with a multitude of sensors, such as LIDARs, cameras, GPS, and various other sensors \cite{its-vehicle,valle2023}.

The IAMCV Dataset was acquired as part of the FWF Austrian Science Fund-funded ``Interaction of Autonomous and Manually-Controlled Vehicles" project. It is primarily centered on inter-vehicle interactions and captures a wide range of road scenes in different locations across Germany, including roundabouts, intersections, and highways. These locations were carefully selected to encompass various traffic scenarios, representative of both urban and rural environments. By simultaneously capturing data from multiple sensor modalities, our dataset provides an in-depth understanding of the road network from a driver-centric perspective, enabling researchers and developers to analyze, model, and evaluate autonomous driving algorithms and systems under diverse conditions.

The primary objective of this study was to introduce this rich dataset and outline its potential applications in advancing the field of autonomous driving. Furthermore, we present a detailed description of the sensor setup, data acquisition methodology, and preprocessing techniques employed to ensure the quality and reliability of the recorded data. The dataset presented here serves as a valuable resource for researchers, engineers, and practitioners working in the fields of autonomous driving, computer vision, and robotics.
\IEEEpubidadjcol

The remainder of this paper is organized as follows: Section~\ref{sec:related_work} provides an overview of datasets pertinent to the advancement of autonomous driving systems. Section~\ref{sec:platform} describes the vehicle used for the data acquisition, sensor arrangement, and calibration procedures. An overview of the IAMCV dataset is presented in Section~\ref{sec:dataset}. Subsequently, the insights of the dataset are analysed in Section~\ref{sec:analysis} including two proof-of-concept applications to showcase the dataset's usability. Final conclusions are provided in Section~\ref{sec:conclusions}.

\section{Related Work}
\label{sec:related_work}

This section provides a summary of the datasets relevant to the progress of autonomous driving systems, focusing on multimodal sensor data. The datasets are presented in a chronological sequence.

According to the authors in \cite{dora_dataset}, datasets can be categorized into three primary types: datasets from fixed stations, drones, and test vehicles. Datasets from fixed stations are derived from sensors (such as cameras, LIDARs, and radars) affixed to traffic lights, poles, or buildings at specific locations, serving dual roles as surveillance systems and sources of research-oriented traffic data. One of the notable datasets in this category is The Next Generation Simulation (NGSIM) Dataset\cite{kovvali2007video}.

Similar to fixed stations, drone datasets offer a stable and location-specific perspective. Notably, datasets of this kind have seen an increasing prevalence in recent times \cite{dora_dataset,ind_dataset,round_dataset,highd_dataset,exid_dataset,interaction_dataset}. Nevertheless, for the development, training, and analysis of perception and control algorithms for ADS-operated vehicles, data from test vehicles remains the primary source of essential information.

In 2013, the KITTI dataset \cite{kitti} was a cornerstone in autonomous vehicle research because of its comprehensive and real-world nature. This dataset encompasses a wide array of sensor data, including high-resolution images, LIDAR point clouds, and GPS information, collected from a vehicle equipped with a variety of sensors, as it navigates urban environments. 

The  FORD Multi-AV dataset \cite{ford_dataset} was collected by a fleet of Ford autonomous vehicles (AVs) on different days from 2017 to 2018. The vehicles traversed an average route of 66 km in Michigan (USA), including freeways, airports, city centers, a university campus, and suburban neighborhoods. The dataset covered seasonal variations in weather, lighting conditions, road occupancy, and traffic experienced in dynamic urban environments over different months. The vehicles were geared with an Applanix POS-LV GNSS/INS system, four HDL-32E Velodyne 3D-lidar scanners, and six Point Grey 1.3 MP cameras arranged to grant a 360° coverage. 

The PREVENTION dataset \cite{prevention_dataset} was introduced in 2019 and offers extensive and precise annotations of vehicle trajectories, categories, lanes, and various events, including cut-in, cut-out, left/right lane changes, and hazardous maneuvers. The dataset was collected by deploying six distinct sensors (LIDAR, radar, and cameras) on an instrumented vehicle operating under naturalistic driving conditions. The PREVENTION dataset encompasses 356 min of data, equivalent to 540 km of travel distance, featuring over 4 million detections and more than 3,000 individual trajectories. 

The EU Longterm dataset \cite{eulongterm_dataset} was gathered in 2020 using the Robocar in the urban and suburban areas of Montbéliard, France. The vehicle speed was maintained at 50 km/h throughout the data collection process. The recorded driving route spans approximately 5.0 kilometers, recorded in 16-minute-long segments. The data collected at 10Hz came from two stereo cameras (pointing back and front) two industrial cameras (facing the sides of the vehicle), two Velodyne HDL-32E LIDARS, one 2D LIDAR, one radar, an IMU, and a GPS receiver, all of which were stored in rosbag files

Further diversification was presented in 2021 with the Waymo Open Dataset \cite{waymo_dataset}. This dataset contains more than 100,000 scenes, each spanning 20 s and captured at a rate of 10 Hz, resulting in more than 570 h of unique data covering a distance of 1750 km along roadways across six cities within the United States. The dataset provides LIDAR scans, camera images, labels, and 3D bounding boxes.

Other widely used datasets include the Oxford RobotCar dataset \cite{RobotCarDatasetIJRR, RadarRobotCarDatasetICRA2020} introduced in 2016 and updated in 2020, nuScenes \cite{nuscenes} introduced in 2020, and ONCE \cite{once_dataset} introduced in 2021. All were recorded with a sensor array, including cameras and LIDARs.

Although existing datasets have proven invaluable, the need for diversification in data sources, sensor configurations, geographical locations, and driving scenarios remains crucial. This diversity is essential for the rigorous training and validation of both new and existing perception and control algorithms. It allows algorithms to generalize better, adapt to novel and complex environments, and enhance their robustness in real-world situations. Moreover, the evolving landscape of autonomous driving, including the introduction of new sensor technologies and the emergence of different driving cultures worldwide, necessitates the continual generation of fresh data. In essence, the production of new datasets serves as the backbone of innovation in autonomous vehicle technology, enabling safer, more efficient, and globally adaptable self-driving systems.

\section{Data Collection}
\label{sec:platform}
This section describes the sensors used in the data acquisition process, the vehicle setup, and the calibration procedures employed to record the dataset.

\subsection{Sensor Overview}
\label{sec:sensors}
This section provides a comprehensive description of the sensors used in the data collection. The dataset was acquired using the JKU-ITS vehicle \cite{its-vehicle}. The vehicle was equipped with different sensors to perceive the surrounding environment and locate the vehicle within it.

\begin{figure}[htbp]
	\centering
	\begin{subfigure}{0.3\textwidth}
		\includegraphics[width=\textwidth]{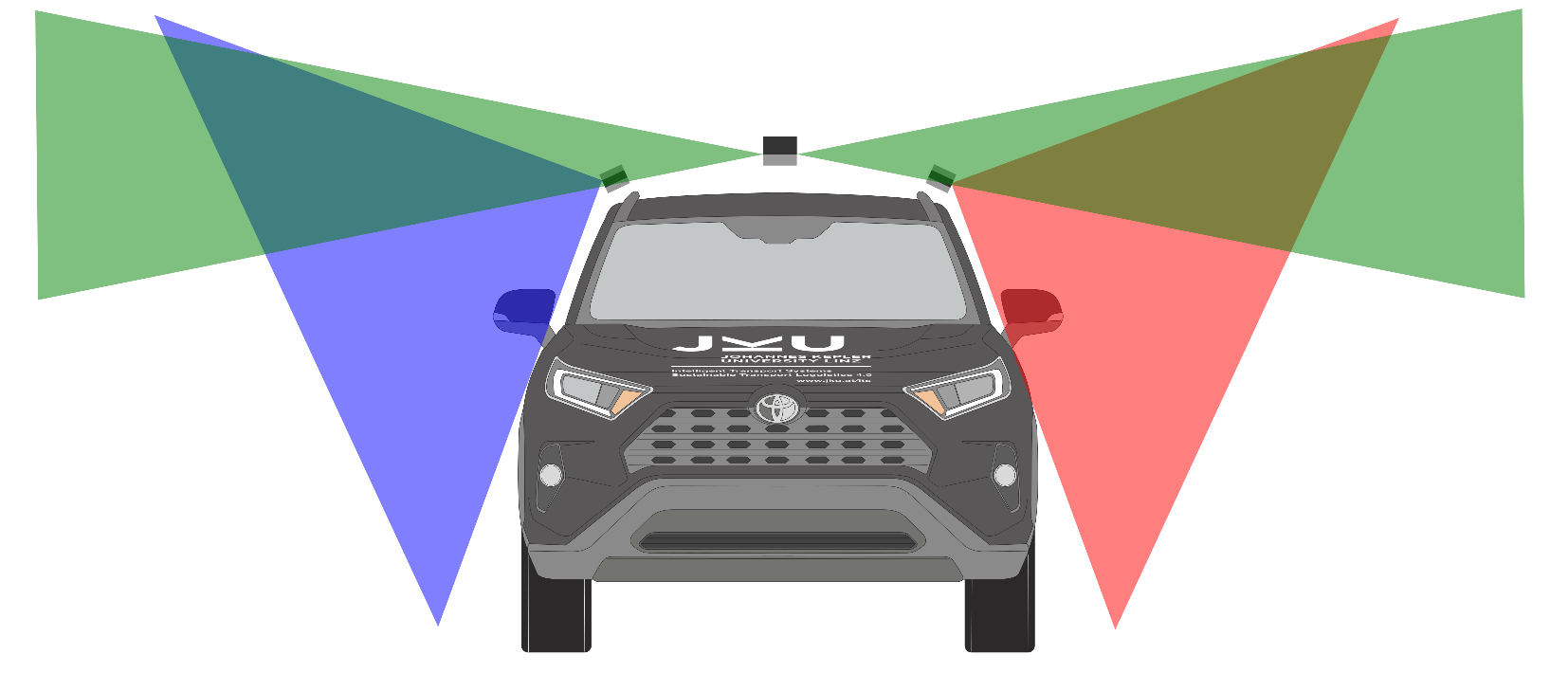}
		\caption{}
	    \label{fig:lvfov}
	\end{subfigure}
	\\
	\begin{subfigure}{0.2\textwidth}
		\includegraphics[width=\textwidth]{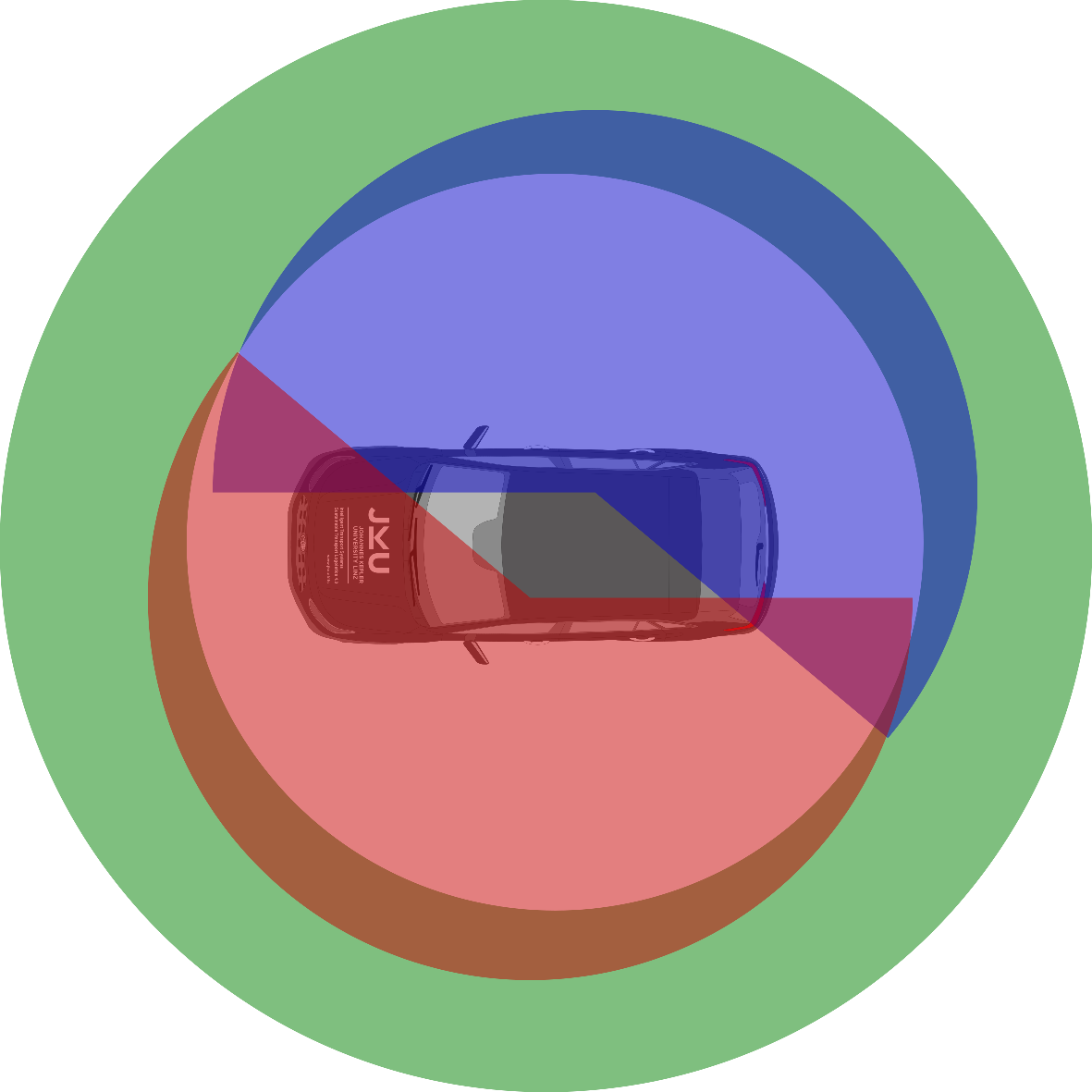}
		\caption{}
	    \label{fig:lhfov}
	\end{subfigure}
    \begin{subfigure}{0.2\textwidth}
		\includegraphics[width=0.65\textwidth]{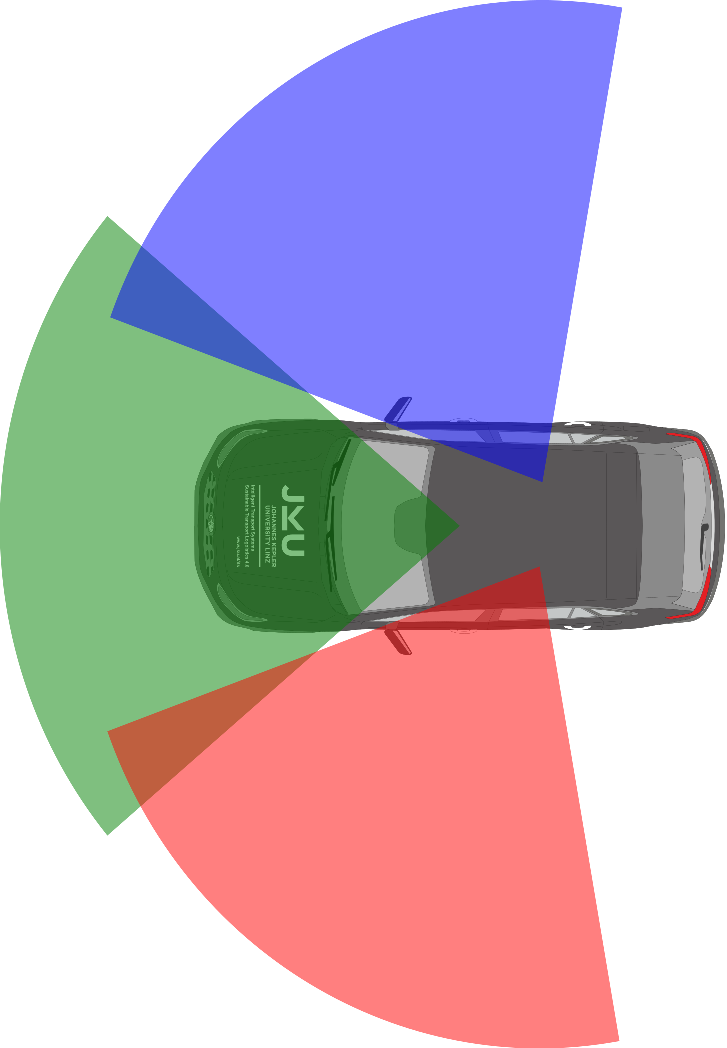}
		\caption{}
	    \label{fig:chfov}
	\end{subfigure}
	\caption{(a) depicts the location of the three \gls{LIDAR} sensors and the vertical \gls{FoV} associated with each sensor. (b) Shows the horizontal \gls{FoV} of the \gls{LIDAR} sensors. Lastly, (c) depicts the horizontal \gls{FoV} covered by the three cameras \cite{its-vehicle}.}
	\label{fig:fov}
\end{figure}

The vehicle was equipped with three \gls{LIDAR}s. An OS2 (64 layers and $22.5^\circ$ of vertical \gls{FoV}) was located in the center of the roof. This sensor has an extended range of view (240m), covering objects far from the vehicle (Fig.~\ref{fig:lvfov}). Two OS0 (128 layers, range from 0.3m to 50m, and $90.0^\circ$ of Vertical \gls{FoV}) detected the objects in the vicinity of the vehicle. One was located in the front-left part of the roof and the other was in the rear-right part. As shown in Fig.~\ref{fig:lvfov}, both OS0 sensors were inclined $22.5^\circ$ towards the plane to cover the sides of the vehicle. This configuration with three \gls{LIDAR} sensors allowed the vehicle to cover the entire area surrounding the car without noticeable dead zones (Fig.~\ref{fig:lhfov}).

\begin{table}[htbp]	
    \centering
    \caption{Sensors' specifications.} 
    \begin{tabular}{|l|c|c|c|c|c|} 
        \hline
        \textbf{Lidars} & \textbf{model} & \textbf{range} & \textbf{VFoV} &  \textbf{HFoV} & \textbf{layers} \\ \hline
        lidar\_c             &    OS2  & 240m   & $22.5^\circ$    & $360^\circ$ & 64 \\ \hline
        lidar\_l &    OS0  & 50m   & $90.0^\circ$    & $360^\circ$ & 128 \\ \hline
        lidar\_r &    OS0  & 50m   & $90.0^\circ$    & $360^\circ$ & 128 \\ \hline
        \textbf{Cameras} & \textbf{model} & - & \textbf{VAoV} &  \textbf{HAoV} & \textbf{resolution} \\ \hline
        camera\_c &   acA2040-35gc  & -   & $59.4^\circ$    & $79.0^\circ$ & 3.5MP \\ \hline
        camera\_l &   acA2040-35gc  & -   & $59.4^\circ$    & $79.0^\circ$ & 3.5MP \\ \hline
        camera\_r &   acA2040-35gc  & -   & $59.4^\circ$    & $79.0^\circ$ & 3.5MP \\ \hline
    \end{tabular}\label{tab:sensors}
\end{table}

\subsubsection{Cameras}
Three acA2040-35gc Basler GigE cameras (3.2MP, Angle of View: $79.0^\circ$ horizontal, $59.4^\circ$ vertical) were located on the roof, covering a horizontal field of view of approximately $200^\circ$, as shown in Fig.~\ref{fig:chfov}. The cameras were configured to acquire 10 frames per second according to an external 10Hz signal (trigger) generated by the synchronization box.

\subsubsection{GNSS/INS}
The positioning system receiver (3DMGX5-45) was located beneath the central \gls{LIDAR} sensor. It offered access to raw data from the \gls{GPS} at a refresh rate of 4Hz, data from the \gls{IMU} at a sample rate of 100Hz, and filtered odometry data obtained through a Kalman filter method at a refresh rate of 100Hz. Furthermore, the GNSS/INS established a direct connection to the main computer through a USB cable.
 
\subsection{Sensor Calibration}

Intrinsic calibration of the cameras was achieved using the standard \gls{ROS} package \textit{camera\_calibration}~\cite{ROS}. This camera calibration model uses the OpenCV library~\cite{opencv} and is based on the pinhole camera model. The model estimates intrinsic camera parameters, such as focal length, principal point coordinates, and lens distortion coefficients. The calibration procedure involves capturing multiple images of a known calibration pattern (chessboard) from different viewpoints, and using the corresponding image and world coordinates to calculate the camera parameters.

The extrinsic calibration of the cameras and \gls{LIDAR} sensors was performed following the automatic procedure described in~\cite{beltran2022} with the available \gls{ROS} package. The three cameras and the two side-\gls{LIDAR}s were calibrated using the center-\gls{LIDAR} location as the reference frame.

\subsection{Vehicle Data Bus}

The OBDII port is a standardized diagnostic connector that allows access to real-time data from the vehicle engine and other systems. The Black Panda is a universal car interface that provides access to the communication buses of the car through this port and a USB-C interface at the other end to connect a computer. A custom bridge based on Python was developed to connect the acquisition software in \gls{ROS}\cite{ROS} with the software interface of the Black Panda (based on the open source system \textit{Openpilot}). Throughout the entirety of this paper, the variables of the vehicle accessed through this method are referred to as ``car state". A list of the most commonly used car state variables is presented in Table~\ref{tab:car_state}. A full list of more than 40 variables can be found in the dataset files \cite{iamcv}.

\begin{table}[ht]	
    \centering
    \caption{car state variables} 
    \begin{tabular}{|p{0.1\textwidth}|p{0.32\textwidth}|} 
        \hline
        \textbf{Name} & \textbf{Description}  \\ \hline
        vEgo            & vehicle linear speed in m/s \\ \hline
        aEgo            & vehicle linear acceleration in $m/s^2$ \\ \hline
        wheelSpeeds     & wheel linear speed in $m/s$ (one measure per wheel) \\ \hline
        gearShifter     & gear level position (park, neutral, drive, etc...) \\ \hline
        steeringAngle   & angle of the steering wheel \\ \hline
        gas             & gas pedal level, where 0 is released and 1 is fully depressed\\ \hline
        gasPressed      & value specifying whether the gas pedal is pressed () or released\\ \hline
        brakePressed    & value specifying whether the brake pedal is pressed or released\\ \hline
        steeringTorque  & torque applied on the steering wheel \\ \hline
    \end{tabular}\label{tab:car_state}
\end{table}

\subsection{Synchronization}
Synchronizing data from different sensors is an important step in creating high-quality datasets for research on autonomous vehicles. Thus, the accuracy and robustness are improved. In the IAMCV dataset, the synchronization was achieved using two methods:
\begin{itemize}
    \item Synchronization by hardware: A hardware box generated six trigger signals for the \gls{LIDAR}s and cameras \cite{its-vehicle}. Through this process, it is guaranteed that the three \gls{LIDAR} beams point to the front of the vehicle simultaneously when the cameras are triggered. However, the timestamps are assigned at the moment the frames are received in the computer running \gls{ROS} and then edited in an offline process to guarantee that for every point cloud, the timestamp corresponds to the moment when the beam was pointing to the front of the vehicle. This is at midscan for the \gls{LIDAR}s in the center and the left and at the beginning of the scan for the one on the right.
    \item Synchronization by software: In the case of the car state variables and the GPS, it was not possible to connect a trigger signal; thus, the data were acquired and the timestamps were assigned at the moment the frames were received in the computer running \gls{ROS}. Some latency between 0ms and 30ms was measured for this procedure.
\end{itemize}

\subsection{Vehicle setup}
\begin{figure}[htbp]
    \centering
    \includegraphics[width=0.48\textwidth]{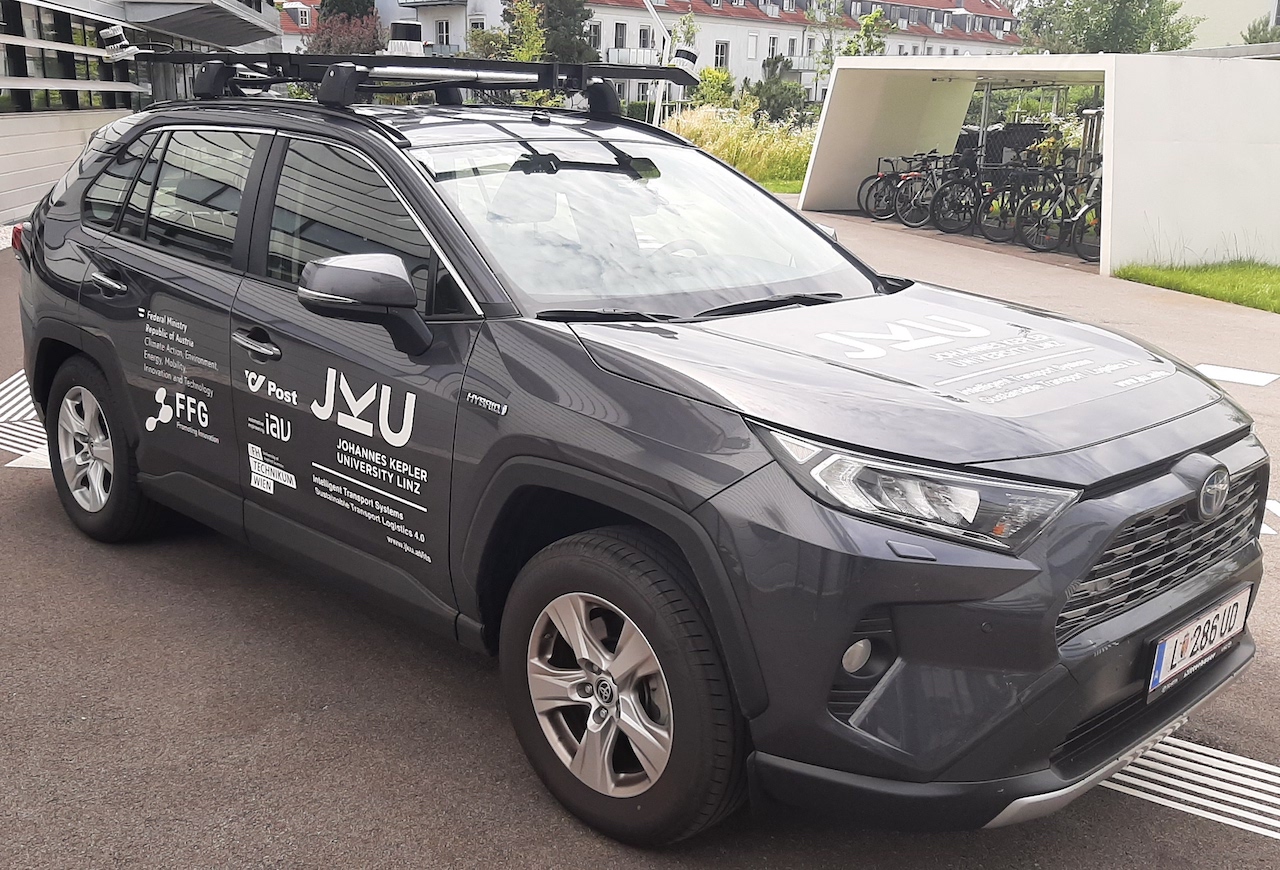}
    \caption{The JKU-ITS vehicle was used for the collection of the data.}
	\label{fig:vehicle}
\end{figure}
 The dataset was acquired using the JKU-ITS vehicle \cite{its-vehicle} as shown in Fig.~\ref{fig:vehicle}. The vehicle was equipped with the sensors described in Section~\ref{sec:sensors}. The cameras and \gls{LIDAR} sensors were connected to the main computer through a network switch. The data acquisition software was based on \gls{ROS}\cite{ROS}. A custom package for launching the \gls{ROS}-driver for each sensor and the associated transformations was developed. The raw data are primarily stored in rosbag files and later converted to more standard options (CSV files, PNG, etc.). 

%%%%%%%%%%%%%%%%%%%%%%%%%%%%%%%%%%%%%%%%%%%%%%%%%%%%%%%%%%%%%%%%%%%%%%%%%%%%%%%%
\section{IAMCV Dataset}
\label{sec:dataset}

\subsection{Methodology}
\label{sec:methodology}
The dataset was recorded from the 3rd to the 10th of June of 2022. The recordings were performed during the daytime, between 08:00 h to 20:00 h under different natural light levels (sunny, cloudy, or overcast). Because the dataset aims to record different interactions between the recording vehicle and other vehicles, trucks, motorcycles, and vulnerable road users, the chosen locations were: intersections, roundabouts, roads, and highways.

Three recording locations for intersections were selected in Aachen, Germany. In suburban areas around the city, two locations were chosen for roundabouts and two for country roads. Finally, highway locations were selected along the A3 Federal Motorway between Passau and Cologne. 
\begin{figure}[htbp]
    \centering
    \includegraphics[width=0.48\textwidth]{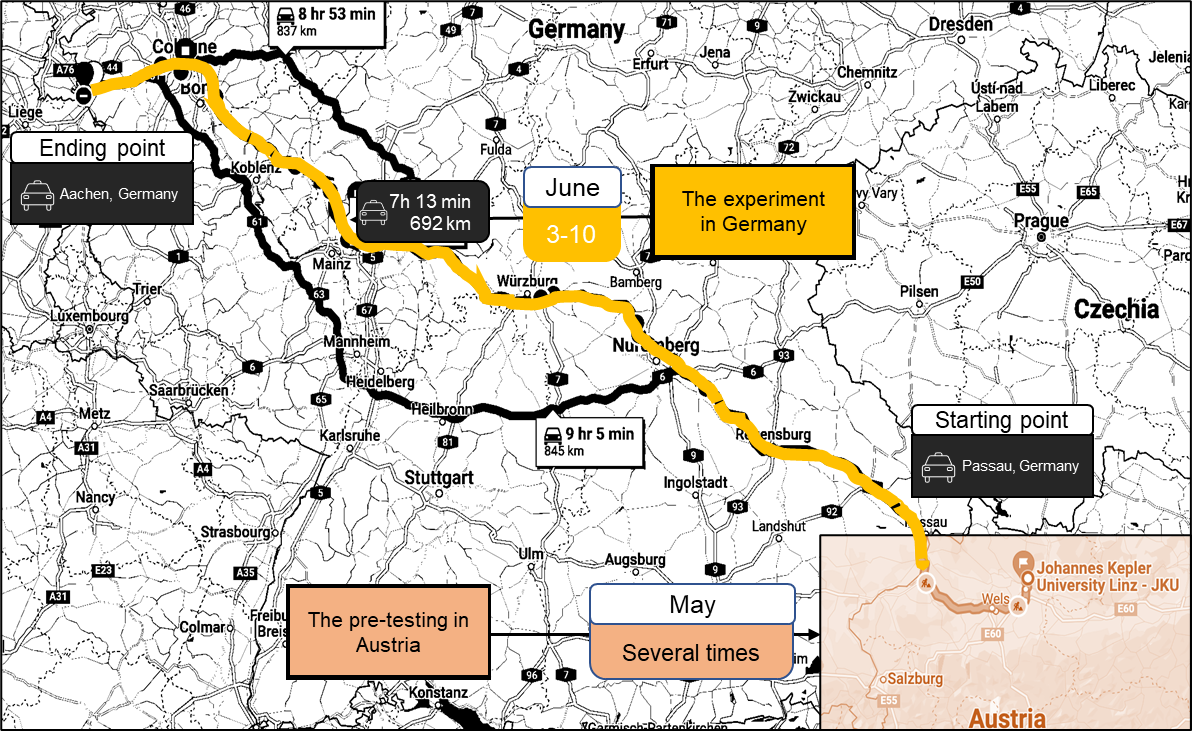}
    \caption{Route followed from Linz (Austria) to Aachen (Germany).}
    \label{fig:dataset_route_map}
\end{figure}
The path in which the data were collected, aligns with the route taken from Linz (Austria) to attend the IEEE Intelligent Vehicles Symposium conference in Aachen (Germany). The entire route is illustrated in Fig. \ref{fig:dataset_route_map}.  It was also taken into consideration that these sites had already been investigated in other datasets using drones \cite{ind_dataset,round_dataset,highd_dataset}. THE IAMCV dataset supplements this approach by introducing driver-centric insights, that enable the analysis of high-level interaction patterns.

The collected data were anticipated to include various driving patterns and trajectories contingent upon specific scenarios and traffic conditions. In the locations selected for intersections and roundabouts analysis, the vehicle traversed the same reference intersection or roundabout several times; thus, the followed path resembles loops around the reference point. These driving patterns are particularly useful for loop-closure purposes when testing mapping or localization algorithms.

\subsection{Data Types} 
The dataset comprises different types of data:
\begin{itemize}
    \item \textbf{Point clouds:} The point clouds from the three \gls{LIDAR}s were stored using \gls{PCD} file format. All original fields that came from the \gls{LIDAR} sensors were also stored in the file:
    \begin{itemize}
        \item x,y,z: coordinates of the target in the reference sensor frame.
        \item Range: distance from the sensor to the target in mm.
        \item Near-infrared: is the ambient level of infrared sensed by the receiver when the transmitter does not point to that area.
        \item Intensity: signal level in the receiver.
        \item Reflectivity: reflectivity level of the target.
        \item Ring: \gls{LIDAR}'s layer. 
    \end{itemize}
    \item \textbf{Images:} Similarly, the images from the three cameras were stored using \gls{PNG} file format. The images were stored without rectification and intrinsic parameters are provided. The images in the dataset were anonymized before publishing by blurring faces and license plates. This was done automatically using scripts based on \cite{CenterFace} and \cite{silva2018a}.
\end{itemize}

Regarding the naming, files from cameras and \gls{LIDAR}s were named using the format ``nn\_sensor\_tttttttttt.tttt.ext" as follows:

\begin{itemize}
    \item nn: stands for the number of recordings. Each recording has a unique two-digit identifier.
    \item sensor: indicates whether the sensor is a camera or a \gls{LIDAR} and its position. e.g., ``camera\_c" stands for the camera located in the center of the vehicle, and ``lidar\_l" stands for the \gls{LIDAR} located in the left side of the vehicle.
    \item tttttttttt.tttt: it corresponds to the timestand of the frame.
    \item ext: file extension .pcd or .png
\end{itemize}

Fig.~\ref{fig:dir_tree} depicts the internal organization of the files that comprised the IAMCV dataset.

\begin{figure}
\begin{minipage}{0.9\textwidth}
\begin{small}
\dirtree{%
.1 <nn>.
.2 camera.
.3 camera\_c.
.4 <nn>\_camera\_c\_<tttttttttt.tttt>.png.
.4 $\ldots$.
.3 camera\_l.
.3 camera\_r.
.2 lidar.
.3 lidar\_c.
.4 <nn>\_lidar\_c\_<tttttttttt.tttt>.pcd.
.4 $\ldots$.
.3 lidar\_l.
.3 lidar\_r.
.2 info.
.3 nn\_nav\_odom.csv.
.3 nn\_carstate.csv.
.3 tf.csv.
.3 camera\_c\_intrinsics.yaml.
.3 camera\_l\_intrinsics.yaml.
.3 camera\_r\_intrinsics.yaml.
.3 $\ldots$.
.2 labels.
}
\end{small}
\end{minipage}
\caption{Directory tree depicting how the folders and files comprising the dataset are organized.}
\label{fig:dir_tree}
\end{figure}

% \subsection{web access}
% \textcolor{red}{ The IAMCV dataset is publicly available on the IEEE Data Port website.}

\section{Dataset Analysis}
\label{sec:analysis}
\begin{figure}[htbp]
	\centering
	\begin{subfigure}{0.15\textwidth}
		\includegraphics[width=\textwidth]{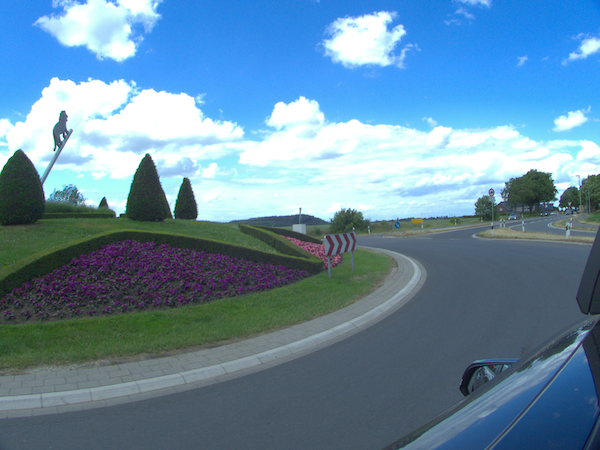}
		\caption{}
	    \label{fig:24_camera_L}
	\end{subfigure}
	\begin{subfigure}{0.15\textwidth}
		\includegraphics[width=\textwidth]{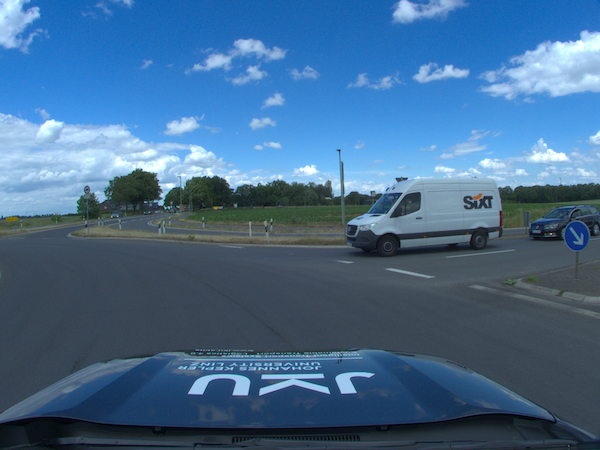}
		\caption{}
	    \label{fig:24_camera_C}
	\end{subfigure}
    \begin{subfigure}{0.15\textwidth}
		\includegraphics[width=\textwidth]{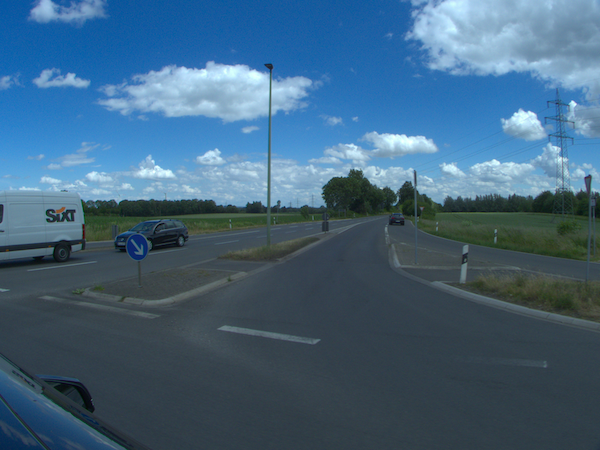}
		\caption{}
	    \label{fig:24_camera_R}
	\end{subfigure}
	\caption{Raw images acquired simultaneously:(a) left camera, (b) center camera, and (c) right camera.}
	\label{fig:24_camera}
\end{figure}

The IAMCV dataset comprises more than 50 segments of 15 min of driving for a total of approximately 15 h of recording. The overall information is presented in Table~\ref{tab:recordings}. Hereunder, some examples of the content of the dataset are discussed. For instance, Fig.~\ref{fig:24_camera} presents a set of three images simultaneously captured by the camera sensors. To facilitate subsequent stitching processes, we ensured a deliberate 20\% overlap between images.
% \begin{table*}[h]	
%     \centering
%     \caption{IAMCV Dataset: Recorded data \textcolor{red}{(There are still blanks in the table)}} 
%     \begin{tabular}{ccccccc} 
%         \toprule
%         location & type & \# & duration &  Date &  Time & distance? \\
%                  &      &    & [min]    & [mm/dd]  & [hh:mm]  \\ \midrule
%         A3 (Germany) & Highway & 8 & 121 & 06/03 & 14:00  \\
%         Frankenberger (Aachen) & Intersection & 4 & 62 & 06/03 & 08:00  \\
%         Bendplatz & Intersection & 3 & 50 & 06/04 & 09:23  \\
%         Frankenberg & Intersection & 5 & 68 & 06/06 & 14:00  \\
%         B57 (Aachen) & Road & 2 & 21 & 06/06 & 14:00  \\
%         Neuweiler & Roundabout & 4 & 72 & 06/06 & 14:00  \\
%         Kackertstrasse & Roundabout & 4 & 66 \\
%         Bendplatz & Intersection & 3 & 50 \\
%         Charlottenburgerallee & Intersection & 4 & 65\\
%         A3 (Germany) & Highway & 19 & 314 \\
%         \textbf{Totals} & - & \textbf{56} & \textbf{888} \\
%         \bottomrule
%     \end{tabular}\label{tab:recordings}
% \end{table*}
\begin{table}[htbp]	
    \centering
    \caption{IAMCV Dataset: Overview of the locations and recording times} 
    \begin{tabular}{|p{0.2\textwidth}|p{0.1\textwidth}|p{0.05\textwidth}|p{0.05\textwidth}|} 
        \hline
        \textbf{location} & \textbf{type} & \textbf{segments} & \textbf{duration} \\
                 &      & \#   & [min]     \\ \hline
        A3 (Germany) & Highway & 27 & 435   \\ \hline
        Aachen city (Frankenberger, Bendplatz, Charlottenburgerallee) & Intersections / Urban& 19 & 295  \\ \hline
        B57 (Aachen) & Country road & 2 & 21 \\ \hline
        Aachen suburbs (Neuweiler, Kackertstrasse) & Roundabout & 8 & 138  \\ \hline
        \textbf{Totals} & - & \textbf{56} & \textbf{888} \\ \hline
    \end{tabular}\label{tab:recordings}
\end{table}

\begin{figure}[htbp]
	\centering
	\begin{subfigure}{0.4\textwidth}
		\includegraphics[width=\textwidth]{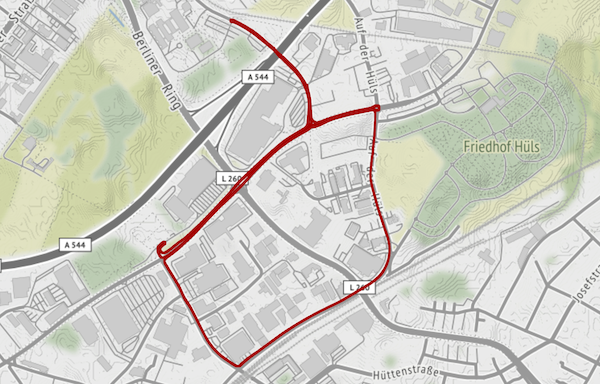}
		\caption{}
	    \label{fig:34_gps_route_map}
	\end{subfigure}
	\begin{subfigure}{0.4\textwidth}
		\includegraphics[width=\textwidth]{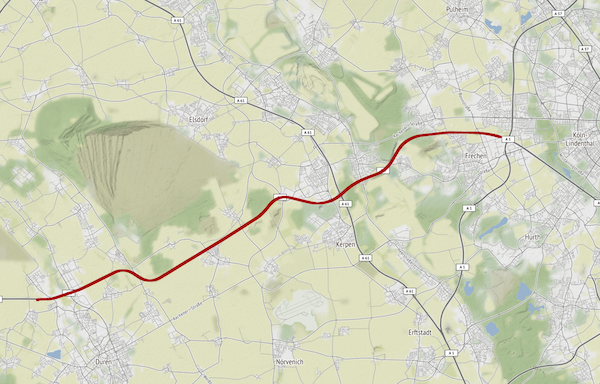}
		\caption{}
	    \label{fig:38_gps_route_map}
	\end{subfigure}
	\caption{Openstreet map with the GPS trajectory followed in two recordings: (a) Recording around Charlottenburgeralle in Aachen. (b) Recording in the A3 highway around Cologne.}
	\label{fig:gps_route_map}
\end{figure}

Fig.~\ref{fig:gps_route_map} depicts the recorded GPS trajectory followed by the vehicle during two recordings. The top image represents a recording that focused on an intersection. As described in Section~\ref{sec:methodology}, the vehicle traversed the same intersection several times. The image at the bottom was recorded on the A3 highway around Cologne.

% \subsection{Discussion}

\begin{figure}[htbp]
    \centering
    \begin{subfigure}{0.22\textwidth}
        \includegraphics[width=\textwidth]{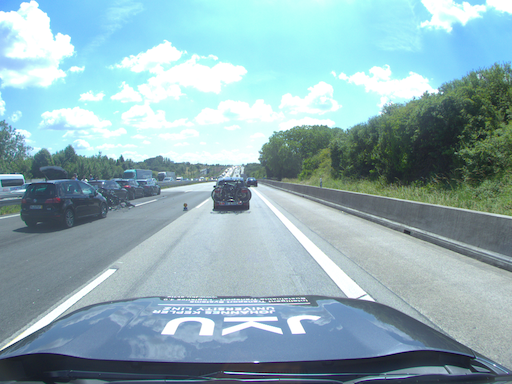}
        \caption{}
        % \label{fig:24_camera_L}
    \end{subfigure}
    \begin{subfigure}{0.22\textwidth}
        \includegraphics[width=\textwidth]{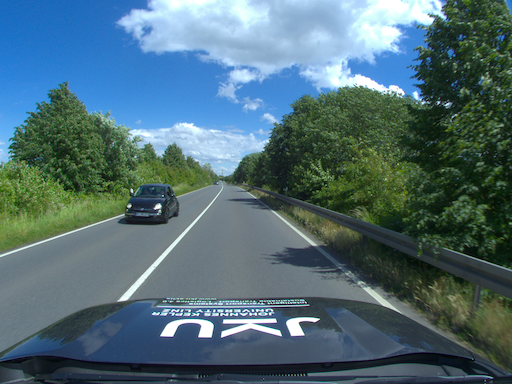}
        \caption{}
        % \label{fig:24_camera_C}
    \end{subfigure}
    \begin{subfigure}{0.22\textwidth}
        \includegraphics[width=\textwidth]{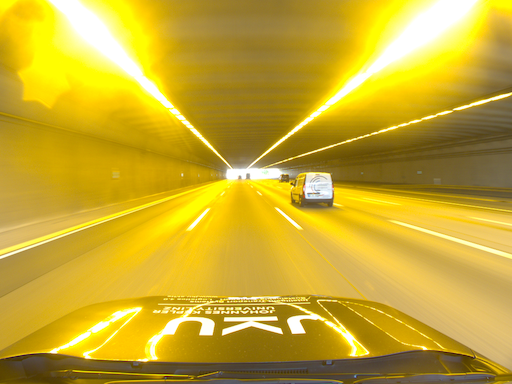}
        \caption{}
        % \label{fig:24_camera_R}
    \end{subfigure}
    \begin{subfigure}{0.22\textwidth}
        \includegraphics[width=\textwidth]{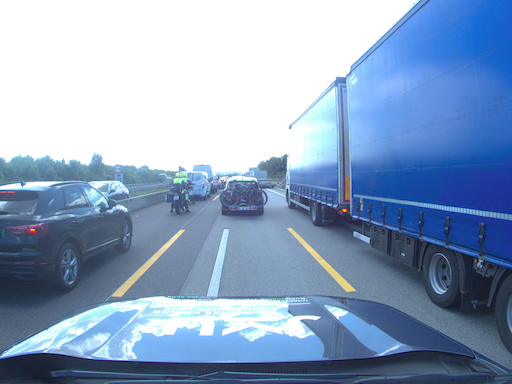}
        \caption{}
        % \label{fig:24_camera_R}
    \end{subfigure}
    \begin{subfigure}{0.22\textwidth}
        \includegraphics[width=\textwidth]{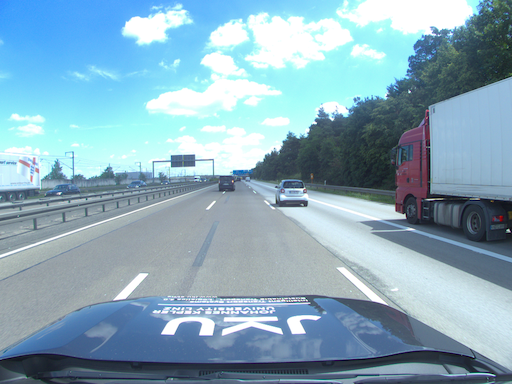}
        \caption{}
        % \label{fig:24_camera_R}
    \end{subfigure}
    \begin{subfigure}{0.22\textwidth}
        \includegraphics[width=\textwidth]{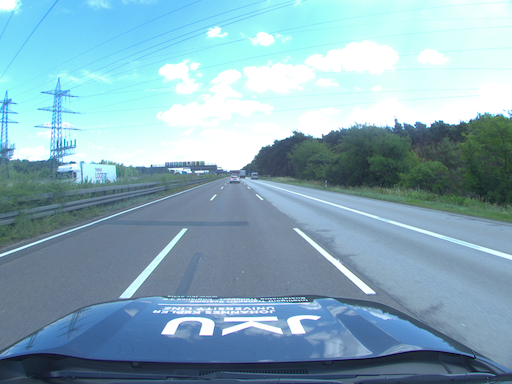}
        \caption{}
        % \label{fig:24_camera_R}
    \end{subfigure}
    \begin{subfigure}{0.22\textwidth}
        \includegraphics[width=\textwidth]{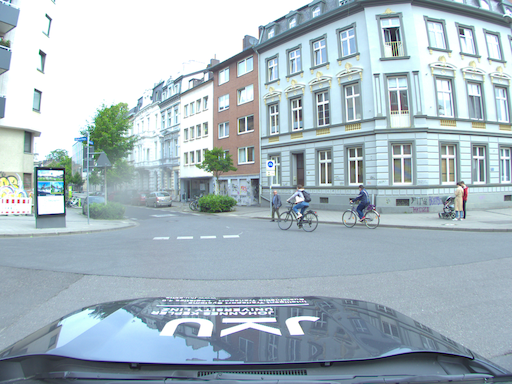}
        \caption{}
        % \label{fig:24_camera_L}
    \end{subfigure}
    \begin{subfigure}{0.22\textwidth}
        \includegraphics[width=\textwidth]{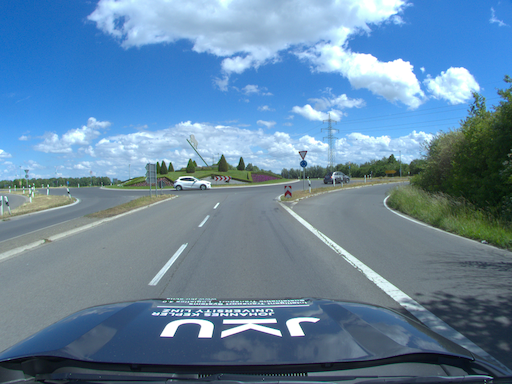}
        \caption{}
        % \label{fig:24_camera_C}
    \end{subfigure}

    \caption{Examples of heterogeneous driving scenes presented within the IAMCV dataset: (a) car crash, (b) country roads, (c) tunnel, (d) traffic jam, (e) overtaking in highway, (f) highway, (g) pedestrian crossing with vulnerable road users, (h) Roundabout.}
    \label{fig:scenarios}
\end{figure}
One of the key advantages of the IAMCV dataset is the large number of heterogeneous scenarios recorded. The availability of these scenarios is a crucial step for testing \gls{ADS}-Operated vehicles and improving the reliability and safety of the algorithms. Six of the most relevant scenarios are depicted in Fig.~\ref{fig:scenarios}; nevertheless, the complete list of available scenarios is considerably larger.

A comparison between the IAMCV dataset and the datasets described in Section~\ref{sec:related_work} is presented in Table~\ref{tab:comparison}. Notably, IAMCV distinguishes itself by incorporating data not only from external GNSS and IMU sources, but also from the vehicle bus itself. This comprehensive approach provides a holistic and fine-grained understanding of vehicle dynamics, sensor interactions, and control inputs, which is essential for developing robust autonomous systems. Furthermore, the IAMCV dataset scenarios and location diversity ensure that algorithms and models trained on the IAMCV dataset are better equipped to handle real-world driving scenarios, making it an invaluable resource for developing and testing autonomous systems. However, it is important to acknowledge that the IAMCV dataset may benefit from further expansion to encompass a broader spectrum of weather conditions including rain, snow, and fog.

\begin{table*}[ht]	
    \centering
    \caption{Comparison between IAMCV Dataset and similar available datasets.} 
    % \begin{tabular}{lllllllll} 
    \begin{tabular}{p{0.18\textwidth}p{0.1\textwidth}p{0.05\textwidth}p{0.04\textwidth}
    p{0.02\textwidth}p{0.07\textwidth}p{0.09\textwidth}p{0.12\textwidth}p{0.12\textwidth}}
\hline
\textbf{Dataset} & \textbf{weather} & \textbf{size} & \textbf{frames} & \textbf{fs}\textsuperscript{1} & \textbf{cameras}\textsuperscript{2}  &  \textbf{LIDAR}\textsuperscript{3}  &  \textbf{vehicle data} & \textbf{location}  \\
 &  &  & \# & [Hz] & pixels &  layers &   &   \\
\hline 
IAMCV (ours) 
& sunny \& cloudy & 7TB  & 530k &10&3x3.5MP &1x64 + 2x128& GPS, IMU, Car state & Urban, highways, country roads  \\
%\hline
KITTI\cite{kitti} 
& sunny \& cloudy &$<$1TB& 15k &10 &4x0.7MP &1x64        & GPS, IMU             & Urban  \\
%\hline
Oxford RobotCar\cite{RobotCarDatasetIJRR, RadarRobotCarDatasetICRA2020} 
& various weather & 23TB & - & 16 & 1x1.2MP + 3x1MP& 1x4 + 2x1 & GPS, IMU & Urban  \\
%\hline
Waymo Open \cite{waymo_dataset}
& various weather& - & 230k & 10 & 5x & 5x & velocity, angular velocity & Urban  \\
%\hline
nuScenes\cite{nuscenes} 
& various weather & 354GB & 400k & 12 & 6x1.4MP & 1x32 & GPS, IMU & Urban  \\
%\hline
A2D2 \cite{a2d2_dataset}
& various weather & 2.3TB & 40k & 10 & 6x2.3MP & 5x16 & GPS, IMU, Car state & Urban, highways, country roads  \\
%\hline
Ford Multi-AV Seasonal \cite{ford_dataset} 
& various weather &$>$10TB & - & 6 & 6x1.3MP + 1.5MP & 4x32 & GPS, IMU & Urban, highways, country roads   \\
%\hline
Prevention \cite{prevention_dataset}
& sunny & 600GB & - & 10 & 2x2MP & 1x32 & GPS, IMU, Car state & Highways \\
%\hline
Eu Long Term \cite{eulongterm_dataset} 
&  various weather  & 870GB & - & 8 & 2 + 2(stereo) & 2x32 + 1x4 + 1x1 & GPS, IMU & Urban  \\
%\hline
eurocity \cite{eurocity_dataset}
& various weather & - & 47k & 0.25 & 1x2MP & No & No & Urban  \\
%\hline
ONCE \cite{once_dataset}
& various weather & - & 1M & 2 & 7 & 1 & No & Urban, highways, country roads  \\
\hline
    \multicolumn{8}{l}{``-" : data not provided.}\\
    \multicolumn{8}{l}{1) fs: sample frequency. In the case of several values, the smaller one is listed.}\\
    \multicolumn{8}{l}{2) AxB where A is the number of cameras and B is the resolution.}\\
    \multicolumn{8}{l}{3) AxB where A is the number of sensors and B is the number of layers in each sensor.}
    \end{tabular}\label{tab:comparison}
\end{table*}

\subsection{Applications}
\label{sec:aaplication}

To demonstrate the IAMCV dataset, serves as a versatile tool with practical applications, two compelling use cases are presented next. In the first instance, we successfully implemented unsupervised trajectory clustering, showcasing the dataset's utility in understanding and categorizing vehicle movements without the need for labeled training data. In a second application, IAMCV proved instrumental in camera calibration. Leveraging the diverse scenarios and sensor data provided by the dataset, we achieved precise calibration, crucial for accurate perception and object localization.

\subsubsection{Unsupervised trajectory segmentation and clustering}

In this application, we tried to segment the recorded trajectories and then cluster the resulting segments into different categories related to the dynamic of the vehicle: braking, turning, accelerating/decelerating. This was accomplished by first segmenting the data using a Bayesian model-based sequence segmentation (BMOSS) algorithm and then clustering the segments by employing an extended latent Dirichlet allocation model (LDA). By investigating line- and trajectory plots, the respective driving action was then derived for each cluster. The method was adapted from \cite{Bender2015}.\\

\begin{figure}[ht]
  \centering \includegraphics[width=0.48\textwidth]{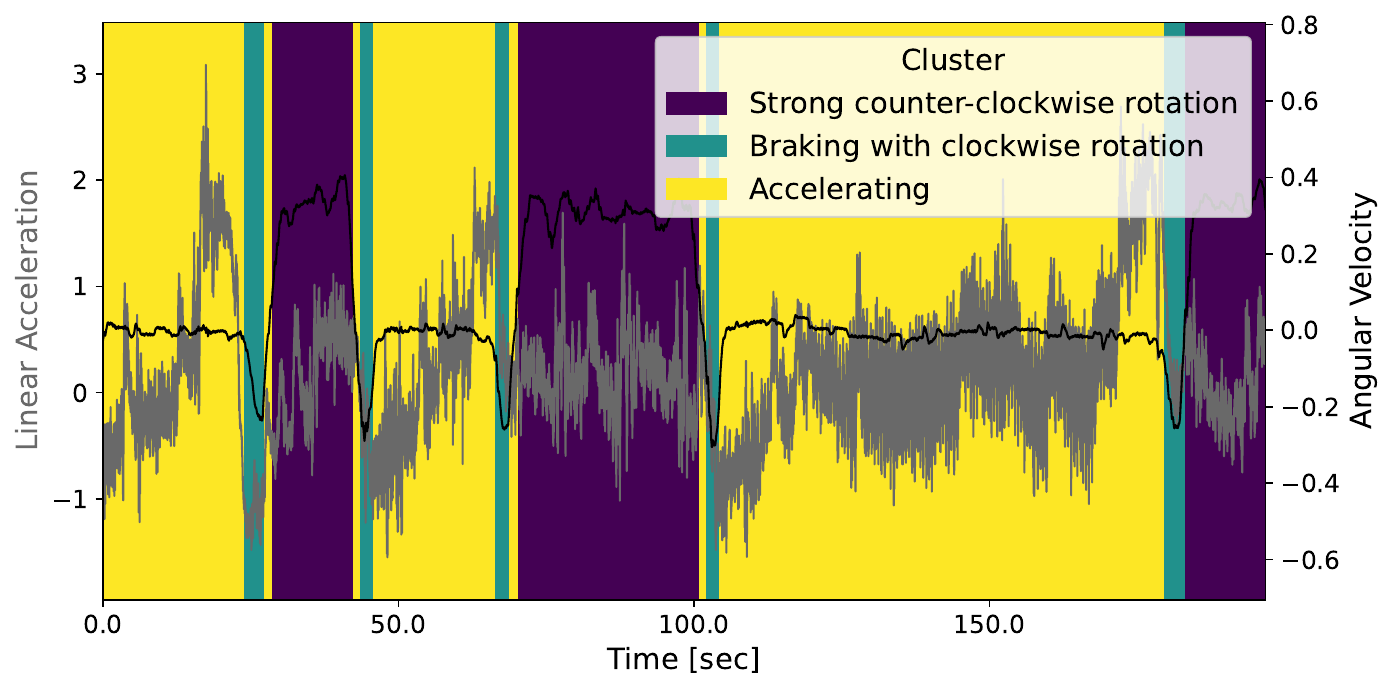}
  \caption{Line plot of the clustered data with the derived driving action for each cluster as a label. }
  \label{fig:line_plot_clustering}
\end{figure}

\begin{figure}[ht]
  \centering \includegraphics[width=0.48\textwidth]{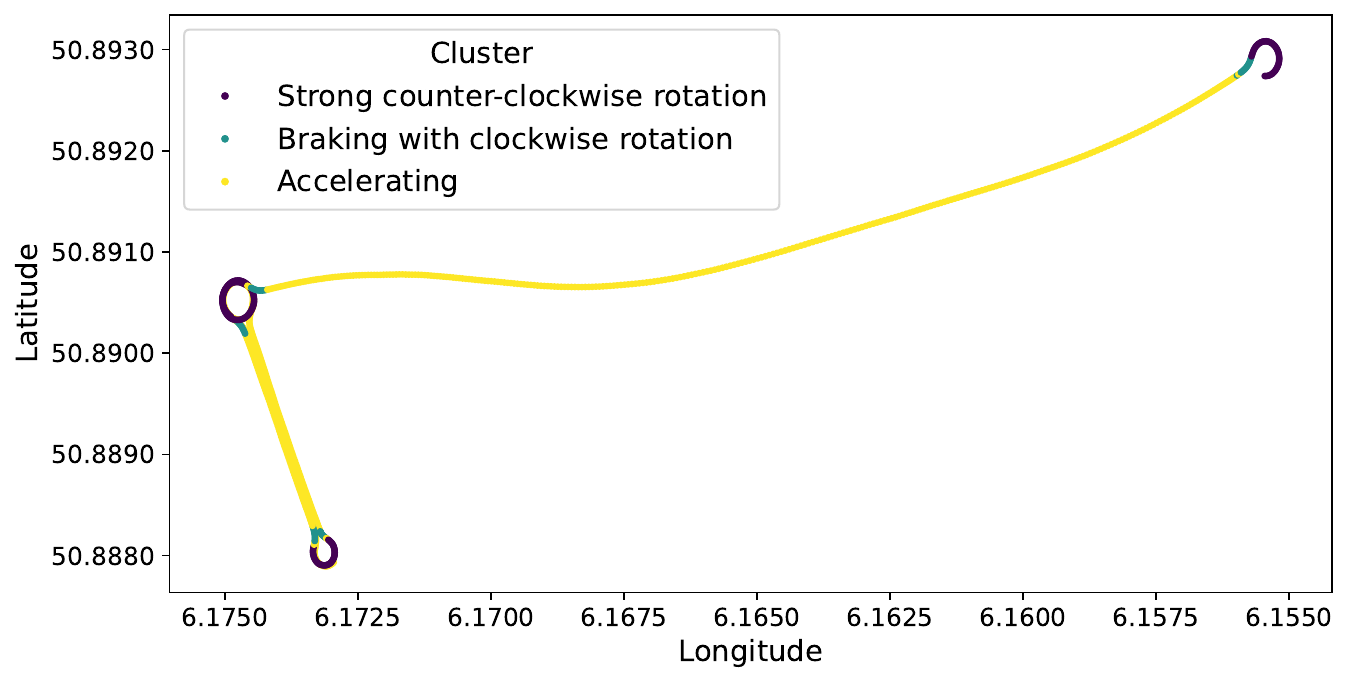}
  \caption{Trajectory plot of the clustered data with the derived driving action for each cluster as a label.}
  \label{fig:trajectory_plot_clustering}
\end{figure}

As parameters linear acceleration ($m/s^2$) along the $x$-axis, i.e. in the direction of motion, and angular velocity ($rad/s$) in the $z$-axis were selected. The data were directly extracted from the IAMCV dataset files without any further preprocessing. With BMOSS, the sequence segmentation problem was formulated as finding the segment boundaries that maximize the overall marginal likelihood of the data. To derive the likelihood of the possible segmentations, the Bayesian piecewise regression model was employed as the underlying framework. It followed the assumption that the sequence of input-output data is piecewise stationary and within each segment the data follow a Gaussian distribution. Traversing the model’s marginal likelihood under all segmentation choices was done recursively with a forward-backward algorithm. The authors in \cite{Bender2015} extend traditional LDA for continuous driving data in a way that instead of inputting the driving samples directly into the model, they are replaced with symbols, where each symbol is a mixture component. Topics, which are typically distributions over discrete input data, are thus represented as mixtures of these symbols. Similar to traditional LDA, the extended model then returns the topic proportions for each segment. The topic with the highest proportion is then assigned to each segment and the driving action related to each topic can be derived. In Fig.~\ref{fig:line_plot_clustering} and Fig.~\ref{fig:trajectory_plot_clustering} the clustering is visualized together with the derived driving actions for one of the recordings performed in roundabouts. It is clear how three distinctive clusters appeared in the trajectory: Strong counter-clockwise rotation (purple), braking with clockwise rotation (dark cyan), and accelerating/decelerating on a straight line (yellow).

\subsubsection{Segment detection for intrinsic camera calibration}
The camera intrinsic matrix is a critical element in computer vision, encapsulating internal parameters like focal length for mapping 3D points to a 2D image plane. Calibration, often involving geometric patterns, establishes these parameters. However, factors such as wear or environmental changes can impact them, necessitating periodic recalibration for precision-demanding applications like robotics or computer vision.

To address this need, we introduced a robust edge segment detection algorithm. Specifically designed for camera calibration, the algorithm enables online calibration without pre-established geometric patterns.

\begin{figure}[htbp]
    \centering
    \begin{subfigure}{0.22\textwidth}
        \includegraphics[width=\textwidth]{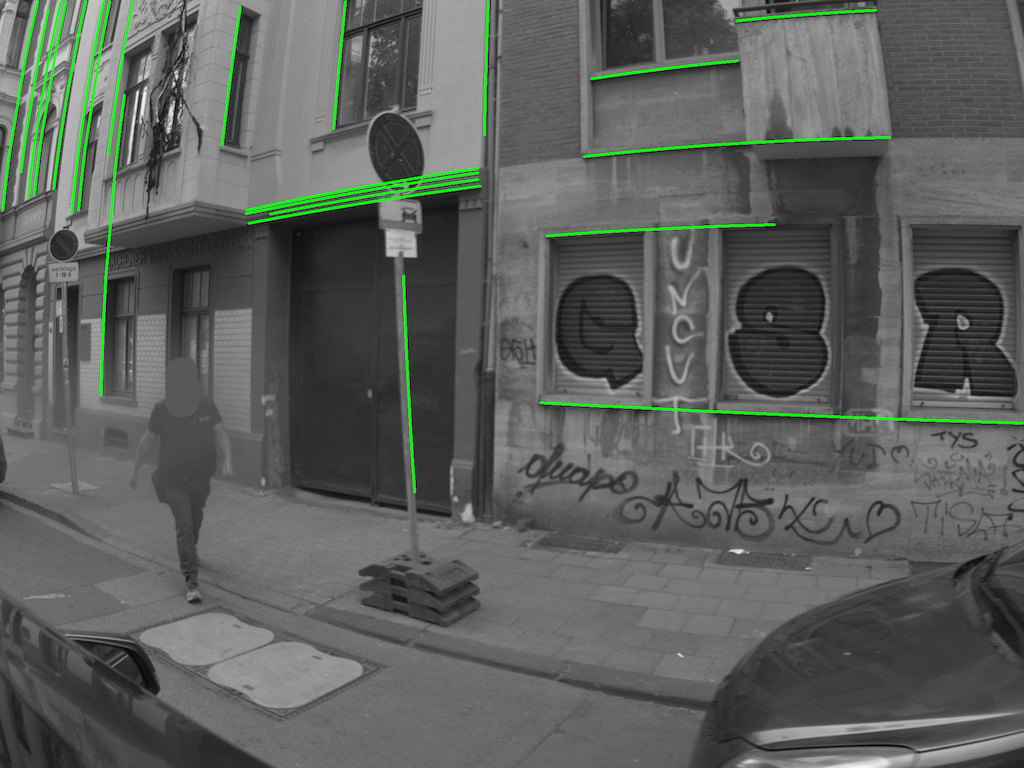}
        \caption{}
        % \label{fig:edges1}
    \end{subfigure}
    \begin{subfigure}{0.22\textwidth}
        \includegraphics[width=\textwidth]{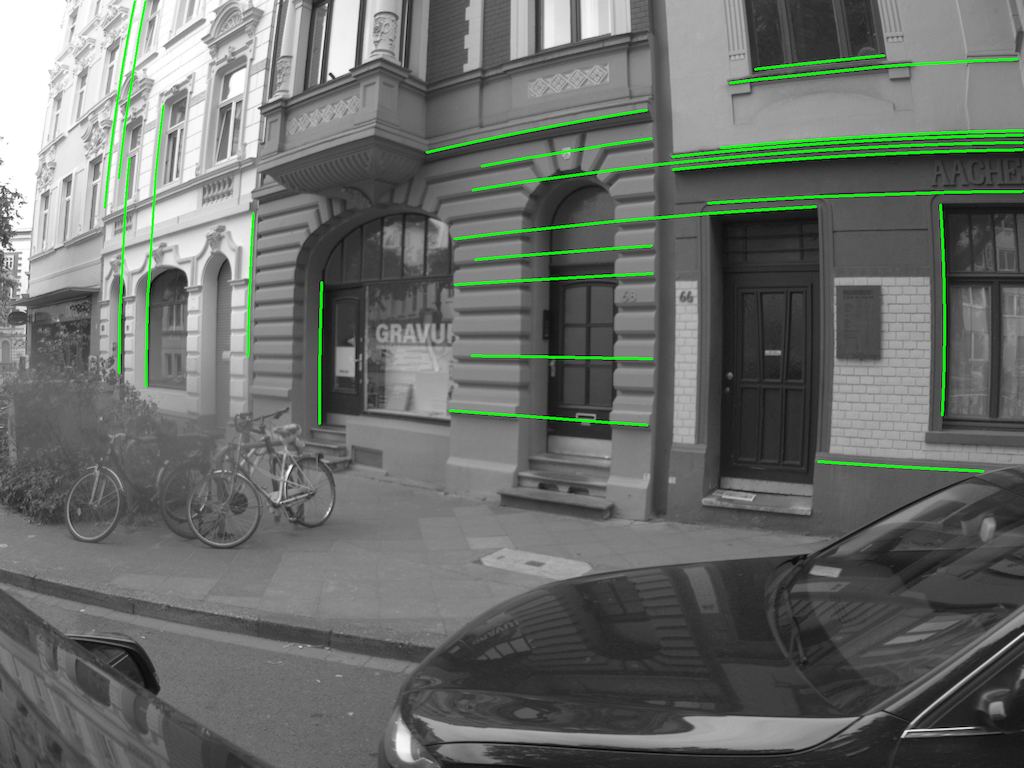}
        \caption{}
        % \label{fig:edges2}
    \end{subfigure}
    \caption{Examples of edge detection on two images extracted from the IAMCV dataset.}
    \label{fig:edge}
\end{figure}

The methodology involves multiple steps for 3D line detection and segmentation from distorted images. First, acceptable orientation ranges are set to prevent the connection of multiple correct edges. Inspired by the Canny algorithm, edge detection follows, including Gaussian filtering, gradient calculation, and thresholding for identifying strong and weak edge points. Weak points are included in the final edge map only if connected to strong points, ensuring edge continuity. Edge pixels are then grouped into continuous segments using a raster scan approach. Shape-based filtering is applied, requiring a minimum extent and aspect ratio for reliable circle fitting. The algorithm addresses 3D straight-line discontinuities by merging associated edge segments, involving circle fitting and residual calculation. Additional shape-based filtering, requiring a minimum length for acceptance, is applied. Finally, outlier filtering eliminates false detections resembling curved 3D lines. This methodology aims to accurately detect and segment 3D lines from distorted images, addressing challenges like noise, discontinuities, and false detections as shown in Fig.~\ref{fig:edge}.

Ultimately, the distortion parameters are extracted from the detected edge segments. In Table~\ref{tab:calibration} the distortion parameters extracted using the IAMCV calibration procedure (ROS-based) are compared with this online approach. Only radial distortion parameters were calculated because their effect is often considered dominant.
\begin{table}[htbp]	
    \centering
    \caption{Comparison of the two main radial distortion parameters estimated using the calibration package in ROS with a chessboard pattern against the online approach using the robust segment detection} 
    \begin{tabular}{|p{0.1\textwidth}|p{0.1\textwidth}|p{0.1\textwidth}|p{0.1\textwidth}} 
         \hline
        \textbf{camera} & \textbf{method} & \textbf{k1} & \textbf{k2} \\ \hline
        left    & ROS    & -0.285356 & 0.082809 \\ \hline
        left    & online & -0.284997 & 0.079993 \\ \hline
        center  & ROS    & -0.283517 & 0.080529 \\ \hline
        center  & online & -0.275294 & 0.073669 \\ \hline
        right   & ROS    & -0.284487 & 0.081967 \\ \hline
        right   & online & -0.274391 & 0.075005 \\ \hline
    \end{tabular}\label{tab:calibration}
\end{table}

\section{Conclusions and Future work}
\label{sec:conclusions}
The IAMCV dataset represents a significant contribution to the field of autonomous driving and sensor-based data collection. Through meticulous data collection and rigorous quality control processes, we have curated a comprehensive and diverse dataset that encapsulates various traffic scenarios across different locations in Germany, including roundabouts, intersections, and highways. This dataset not only serves as a valuable resource for validating object detection, tracking, 3D mapping, and localization algorithms to advance autonomous vehicle technology but also opens doors to innovative research opportunities in sensor fusion, machine learning, and real-world transportation applications. As we look ahead, we envision this dataset becoming a cornerstone for researchers and engineers working towards safer, more efficient, and more intelligent transportation systems. Future iterations of the dataset will take into account weather conditions to further enhance its applicability. 

Two distinct use cases were presented to underscore the dataset's adaptability and effectiveness in addressing multiple challenges within the realm of autonomous systems. As we continue to explore IAMCV's potential, it becomes increasingly evident that its comprehensive and varied data make it an invaluable resource for a spectrum of applications, further solidifying its significance in advancing the field of autonomous driving research.

We plan to include annotations in the form of bounding boxes and high-level labels for selected segments, highlighting relevant interactions in the next release. These real-world data and annotations will play a vital role in advancing the development of autonomous driving systems, making them more robust and dependable. Ultimately, this progress will contribute to the enhancement of safety and efficiency in the field of autonomous vehicles. 

\section*{Acknowledgment}

This work was supported by the Austrian Science Fund (FWF), project number P 34485-N. It was additionally supported by the Austrian Ministry for Climate Action, Environment, Energy, Mobility, Innovation, and Technology (BMK) Endowed Professorship for Sustainable Transport Logistics 4.0., IAV France S.A.S.U., IAV GmbH, Austrian Post AG and the UAS Technikum Wien.

\bibliographystyle{IEEEtran}
\bibliography{paper}
\begin{IEEEbiography}[{\includegraphics[width=1in,height=1.25in,clip,keepaspectratio]{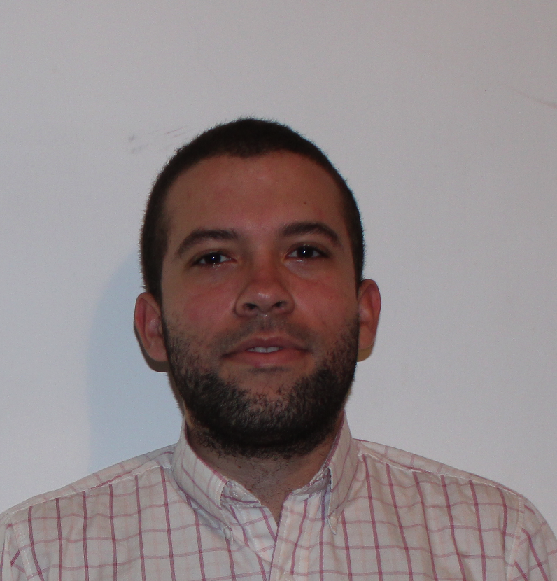}}]{Novel Certad} holds a Master of Science in Electronic Engineering from Simon Bolivar University (USB) in Caracas, Venezuela, awarded in June 2013. As a research assistant, he is currently pursuing a Ph.D. at the Department of Intelligent Transport Systems, at the Johannes Kepler University Linz, Austria. His research areas encompass autonomous vehicles, interactions with road users, signal processing, and probabilistic robotics.
\end{IEEEbiography}

\begin{IEEEbiography}[{\includegraphics[width=1in,height=1.25in,clip,keepaspectratio]{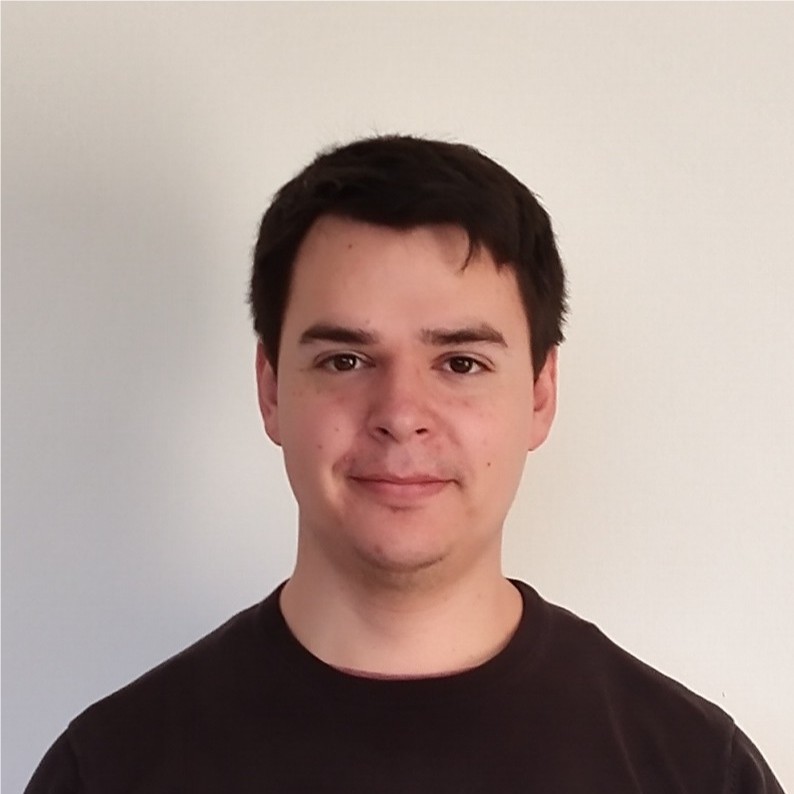}}]{Enrico Del Re} graduated with a Master of Science in Physics 2019 from ETH Zürich in Switzerland and is currently pursuing a Ph.D. at the Department of Intelligent Transport Systems at the Johannes Kepler University Linz, Austria. His research there is focused on the safety of autonomous vehicles, especially in mixed-traffic situations.
\end{IEEEbiography}

\begin{IEEEbiography}[{\includegraphics[width=1in,height=1.25in,clip,keepaspectratio]{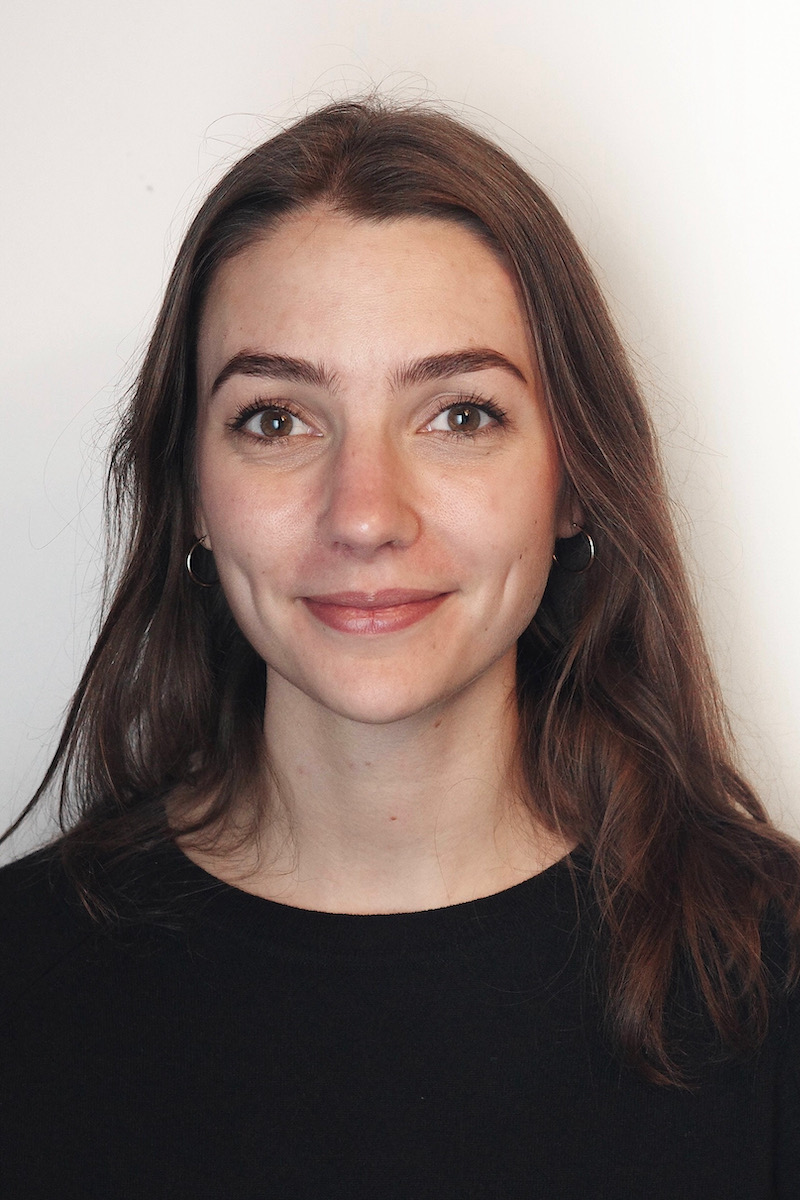}}]{Helena Korndoerfer} 
received a Bachelor of Arts in Economics \& Politics from the University of Munster, Germany in 2020. She is currently pursuing a Master of Science degree in Economic \& Business Analytics at the Johannes Kepler University Linz, Austria and is writing her Master's thesis at the Department of Intelligent Transport Systems.
\end{IEEEbiography}

\begin{IEEEbiography}[{\includegraphics[width=1in,height=1.25in,clip,keepaspectratio]{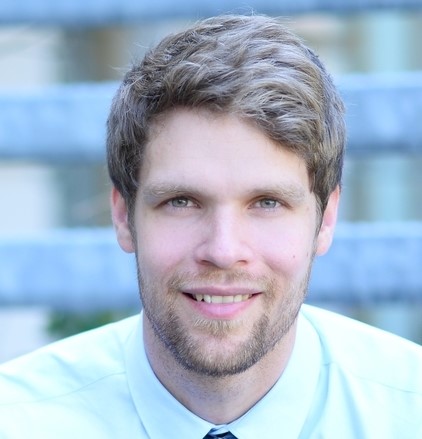}}]{Gregory Schroeder} 
received his Master of Science degree in Computational Engineering Science from the Technische Universität Berlin, Germany, in 2018. 
Since then, he has been engaged in the automotive industry, focusing on pre-development projects in assisted and autonomous driving technologies.
He specializes in Computer Vision, visual Simultaneous Localization and Mapping (SLAM), and camera models \& calibration. His work emphasizes real-world applications, particularly in developing sophisticated and reliable visual systems. His efforts are instrumental in bridging the gap between theoretical computational concepts and their practical implementation in automotive settings.
\end{IEEEbiography}

\begin{IEEEbiography}[{\includegraphics[width=1in,height=1.25in,clip,keepaspectratio]{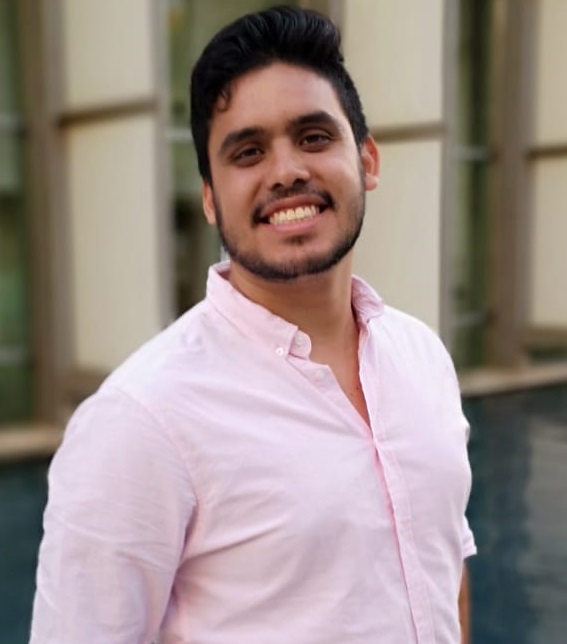}}]{Walter Morales-Alvarez} graduated in 2018 from the Simon Bolivar University in Caracas as an Electronics Engineer. He is a PhD student at Johannes Kepler University, where he is working full-time as a university assistant/lecturer in the Department of Intelligent Transport Systems. His research focuses on the field of Intelligent Transport Systems. He has published in conferences such as the IEEE International Conference on Intelligent Transport Systems and the IEEE Intelligent Vehicle Symposium. His research interests include autonomous driving human-machine interaction and deep learning.
\end{IEEEbiography}

\begin{IEEEbiography}[{\includegraphics[width=1in,height=1.25in,clip,keepaspectratio]{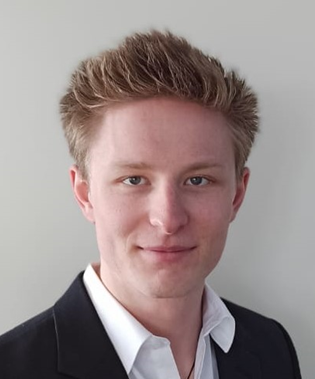}}]{Sebastian Tschernuth} received a bachelor's degree in Biological Chemistry at the JKU Linz and the USB Budweis, and currently studies Artificial Intelligence Masters at the JKU Linz. As a project researcher at the Department of Intelligent Transport Systems, he works on Data Analysis and Machine Learning.
\end{IEEEbiography}

\begin{IEEEbiography}[{\includegraphics[width=1in,height=1.25in,clip,keepaspectratio]{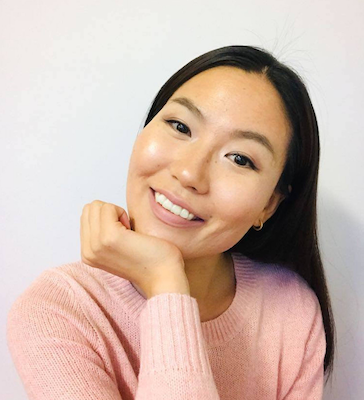}}]{Delgermaa Gankhuyag} received a bachelor’s degree in statistics and economics from the National University of Mongolia, Ulaanbaatar, Mongolia, in June 2012. She is currently pursuing a master's degree in statistics at Johannes Kepler University, Linz, Austria. She is a project researcher at the Department of Intelligent Transport Systems, Johannes Kepler University, Linz, Austria. Her current research interest includes autonomous vehicles and road users interaction, autonomous vehicle’s trajectory prediction, and naturalistic driving behavior of drivers at unsignalized intersections.
\end{IEEEbiography}

\begin{IEEEbiography}[{\includegraphics[width=1in,height=1.25in,clip,keepaspectratio]{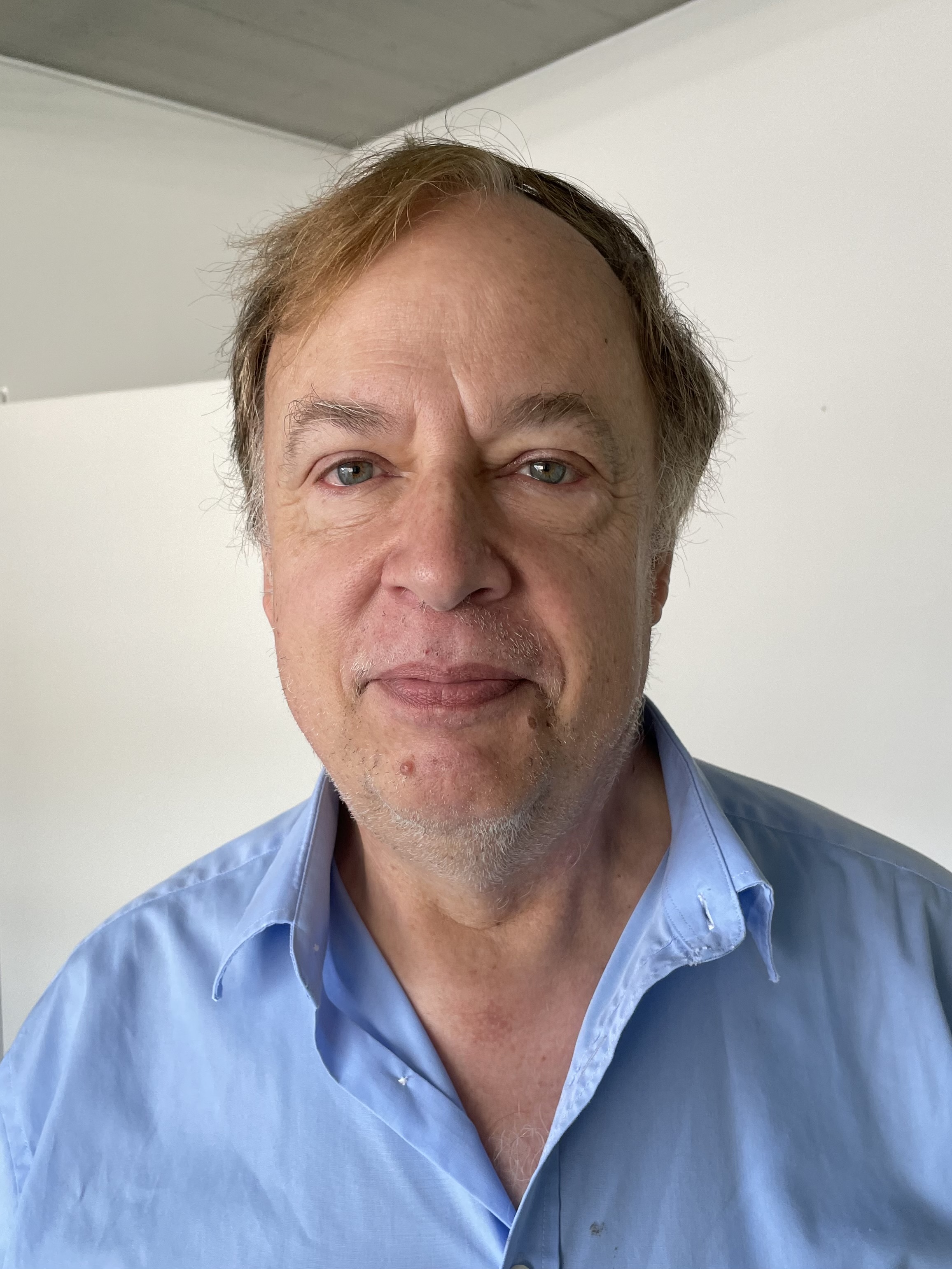}}]{Luigi del Re} (SM'11) received the Doctoral degree in electrical engineering from the Department of Mechanical Engineering, ETH, Zurich, Switzerland. He was a Research Associate with the Institute of Automatic Control, ETH. He left the ETH in 1994 to work for the Swatch group on different automotive issues and he has been a Full Professor with Johannes Kepler University Linz, Austria, since 1998. His main applications are in the field of engine and vehicle technology, hydraulics, process control, and biomedicine. His current research interests include complex control problems which cannot be solved using purely analytical approaches, and, in this context, both nonlinear control design and nonlinear identification, in particular using approximation classes.
\end{IEEEbiography}

\begin{IEEEbiography}[{\includegraphics[width=1in,height=1.25in,clip,keepaspectratio]{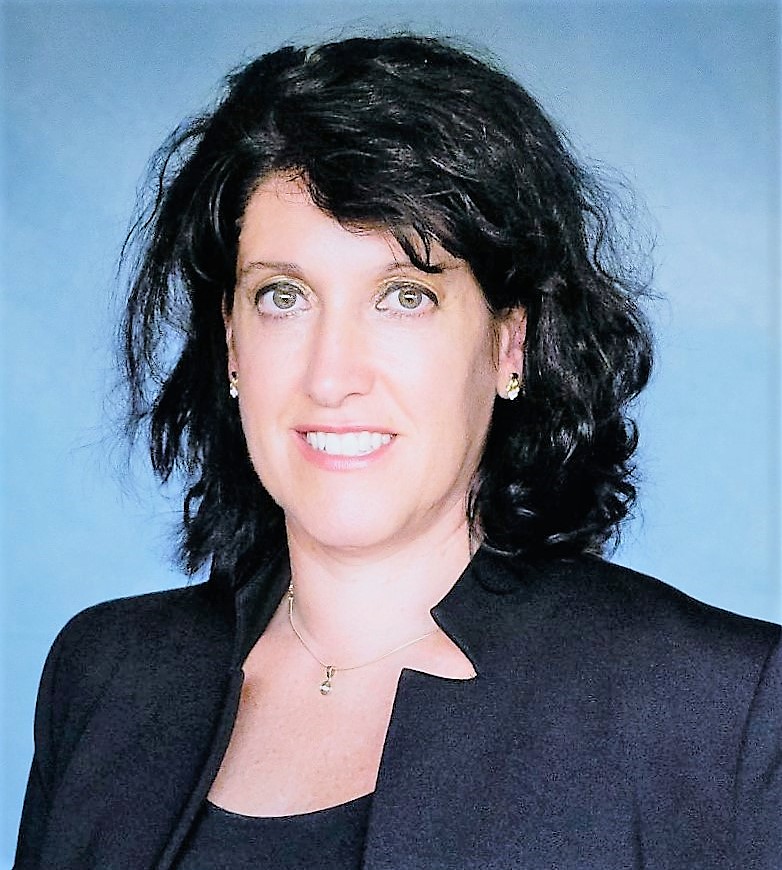}}]{Cristina Olaverri-Monreal} is professor and head of the Department of Intelligent Transport Systems at the Johannes Kepler University Linz, in Austria. Prior to this position, she led diverse teams in the industry and in the academia in the US and in distinct countries in Europe. 

She is also the president of the IEEE Intelligent Transportation Systems Society (IEEE ITSS), founder and chair of the Austrian IEEE ITSS chapter, and chair of the Technical Activities Committee (TAC) on Human Factors in ITS.

She received her PhD from the Ludwig-Maximilians University (LMU) in Munich in cooperation with BMW. Her research aims at studying solutions for efficient and effective transportation focusing on minimizing the barrier between users and road systems. To this end, she relies on the automation, wireless communication, and sensing technologies that pertain to the field of Intelligent Transportation Systems (ITS). 

Prof. Olaverri is a member of the EU-wide platform for coordinating open road tests (Cooperative, Connected and Automated Mobility (CCAM)) as well as a representative for the European technology platform "Alliance for Logistics Innovation through Collaboration in Europe" (ALICE) for the "Workgroup Road Safety" (WG4: EU-CCAM-WG-ROAD-SAFETY@ec.europa.eu). She is additionally a senior/associate editor and editorial board member of several journals in the field, including the IEEE ITS Transactions and IEEE ITS Magazine.

Furthermore, she is an expert for the European Commission on "Automated Road Transport" and consultant and project evaluator in the field of ICT and "Connected, Cooperative Autonomous Mobility Systems" for various EU and national agencies as well as organizations in Germany, Sweden, France, Ireland, etc. In 2017, she was the general chair of the "IEEE International Conference on Vehicles Electronics and Safety" (ICVES'2017). She was awarded the "IEEE Educational Activities Board Meritorious Achievement Award in Continuing Education" for her dedicated contribution to continuing education in the field of ITS.
 
\end{IEEEbiography}

\vfill

\end{document}